%% file: root.tex
\documentclass[letterpaper, 10 pt, conference]{ieeeconf}  

\IEEEoverridecommandlockouts                              

\usepackage{balance}
\input{ieeepreamble}

\title{\LARGE \bf%
Finger Grip Force Estimation from Video using Two Stream Approach
}

\author{Andrey Sartison$^{1, 3}$,
Dima Mironov$^{1}$,
Kamal Youcef-Toumi$^{2}$,
Dzmitry Tsetserukou$^{1}$
\thanks{*This work was supported by Skoltech NGP}
\thanks{$^{1}$The authors are with Space Center, Skolkovo Institute of Science and Technology (Skoltech), 127055 Moscow, Russia}%
\thanks{$^{2}$The author is with Mechatronics Research Laboratory, Massachusetts Institute of Technology (MIT), Cambridge, USA}%
\thanks{$^{3}${\tt\small andrey.sartison@skoltech.ru}}%
}

\begin{document}

\maketitle
\thispagestyle{empty}
\pagestyle{empty}

\begin{abstract}
Estimation of a hand grip force is essential for the understanding of force pattern during the execution of assembly or disassembly operations. Human demonstration of a correct way of doing an operation is a powerful source of information which can be used for guided robot teaching. Typically to assess this problem instrumented approach is used, which requires hand or object mounted devices and poses an inconvenience for an operator or limits the scope of addressable objects. The work demonstrates that contact force may be estimated using a noninvasive contactless method with the help of vision system alone.

We propose a two-stream approach for video processing, which utilizes both spatial information of each frame and dynamic information of frame change. In this work, image processing and machine learning techniques are used along with dense optical flow for frame change tracking and Kalman filter is used for stream fusion.

Our studies show that the proposed method can successfully estimate contact grip force with RMSE $< 10\%$ of sensor range (RMSE $\approx 0.2$ N), the performances of each stream and overall method performance are reported. The proposed method has a wide range of applications, including robot teaching through demonstration, haptic force feedback, and validation of human-performed operations.

\end{abstract}

\section{Introduction}
\label{sec:Introduction}

Automated disassembly is one of the enabling technologies for sustainable development because it is a clean and energy-efficient way to recover valuable materials from e-waste. Limited landfill capacity, unstable material prices, and growing labor costs make automation feasible and raise interest in the field, which can be demonstrated by the development of Liam \cite{liam}, an automated line for iPhone 6 thorough disassembly. It consists of more than 20 tailored robots, which step by step executes the disassembly sequence. However, this approach is principally limited only to large-scale mass-produced products. Our goal in RecyBot, a joint Skoltech-MIT project, is to develop a universal high-speed intelligent robotic system for electronics recycling. RecyBot system consists of several robots, each tailored to perform a specific subset of tasks, whose joined target is to disassemble smartphones to the component level and enable material recovery.


With the current presence of collaborative robots, force feedback is increasingly commonly used in many automation applications, such as gentle object handling and insertion operations with tight margins, where active compliance is essential. In RecyBot these include, for example, screen disassembly, clamp removal, manipulations with fragile and flexible components, and cable disconnection. A significant number of their varieties encourages usage of machine learning methods to deduce the position and the compliance of the end effectors for the operations, but this approach is limited by the absence of the relevant dataset. 

To create such a dataset an invasive instrumented approach may be used \cite{mascaro2001finger,mascaro2004measurement,sun2009estimation}, when disassembly tools are equipped with a force sensor and markers and the procedure of smartphone disassembly by a skilled human with this tools is recorded. The data, collected using this approach usually has a small statistical error, but big efforts are required to collect it and systematic errors may be present due to invasiveness or other restrictions, like the ability measure force only at specific points of manipulated object \cite{pham2017hand}. Other studies require markers to be applied to a hand or a part of a hand \cite{sun2008predicting,urban2013computing}.

On the other hand, noninvasive approach, when position and force are estimated from videos strongly benefits from the vast number of smartphone disassembly videos available online. The position estimation in this approach is well studied \cite{erol2007vision, wei2016convolutional}. State of the art algorithm \cite{wei2016convolutional} uses deep convolutional architecture combined with the implicit spatial modeling.
In the same time, gripping force measurements remain a challenging task. It is a combination of dynamic forces induced by motion of hand with an object and force, caused by fingers pressure pattern itself. In our work, it is assumed that dynamic forces\footnote{See \cite{pham2017hand} for details.} are negligible as our interaction object remains fixed.
 
To fully exploit recent advance of robots learning by showing \cite{maeda2002teaching, zhao2016intuitive}, we propose to add force profile measurement to existing hand-tracking systems. Current studies describe methods of obtaining information on fingers position during the grasp \cite{erol2007vision, wei2016convolutional, rautaray2015vision}, but a very limited set of works emphasize on actual force information. A prominent class of works is based on the property of human fingernail to change its color when force is applied to a tip \cite{sun2009estimation,urban2013computing,chen2014estimating}. In \cite{chen2014estimating} authors used Convolutional Neural Network (CNN) to find a finger alignment and texture mapping to produce a properly aligned representation of a fingernail. Gaussian Process based on aligned images was used to estimate force and torque applied to a fingertip. This method demonstrated good results. However, it has strict limitations on camera position: the camera should always face a nail.
 
In our approach, only video information is used to estimate gripping force at any interaction point. It reduces limitations on camera position relative to hand in comparison with \cite{sun2009estimation}, \cite{sun2008predicting},\cite{pham2017hand} and utilizes dynamic information of hand movement to enhance the precision.
 
In this work, we will concentrate only on normal gripping force estimation of a two fingers grip. The proposed algorithm is a two-stream approach, with a temporal and a spatial stream. Firstly, this paper describes a hardware setup for collecting a dataset. Then we cover algorithm implementation details: signal filtering, image preprocessing techniques, machine learning model and data fusion. In the end, we provide algorithm’s work results.

\section{Methodology}
\label{sec:methods}
In this section, we describe the hardware setup used to acquire the data and the approach used to process that data to estimate gripping force.

\subsection{Hardware setup}
\label{subsec:hardware}
\begin{figure}[h]
    \centering
    \includegraphics[width=0.8\linewidth]{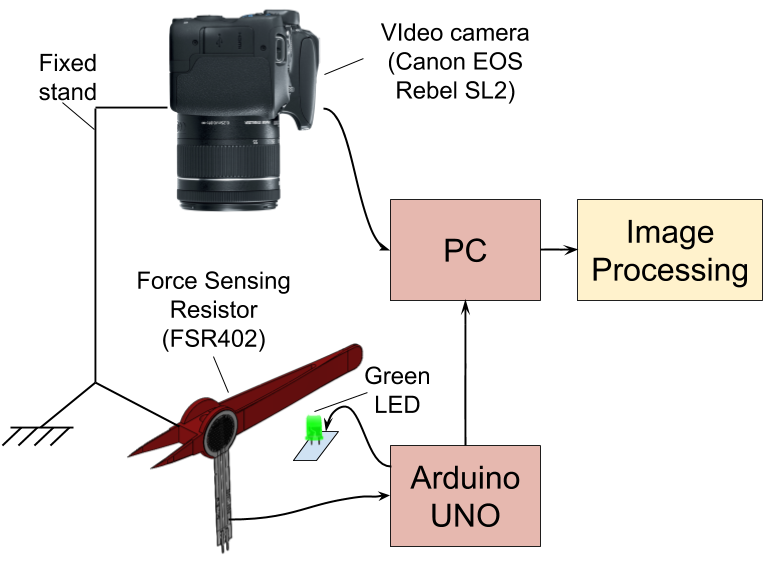}
    \caption{Hardware setup}
    \label{fig:hardware}
\end{figure}

The video and force measurements recording setup is shown on Fig. \ref{fig:hardware}. 
The special stand keeps the camera stationary, and rigid tweezers model fixed relatively to the camera. The rigid tweezers model was 3D printed. It also includes a round marker of distinguishing yellow color. Canon EOS Rebel SL2 camera is used for shooting videos. Videos have Full HD (1920x1080) resolution and 59.95 fps frame rate. A pair of Force Sensing Resistors FSR 402 by Interlink Electronics is used as a force sensor. Green LED is used to synchronize video and force measurements: it lights up when measurements recording starts and goes down after 100 seconds. Arduino UNO is used for analog to digital conversion of FSR signals and transmission of the signal to PC for further processing. The frequency of measurements reading is 100 Hz. Each measurement is presented in the form of two raw 10 bits’ measurement (for left and right sensor respectively) with millisecond timestamp. 

\subsection{Two stream architecture for force estimation}
\label{subsec:architecture}
\begin{figure}[h]
    \centering
    \includegraphics[width=0.9\linewidth]{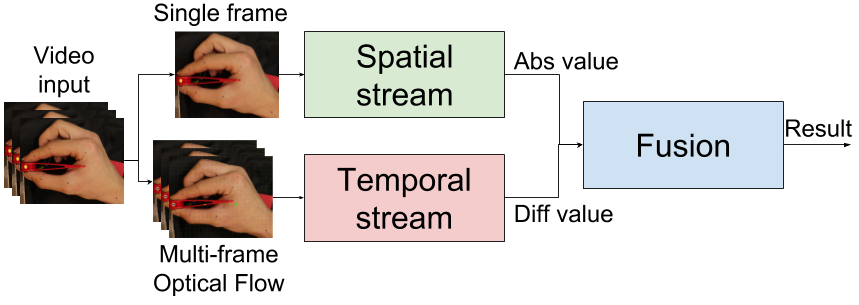}
    \caption{Two stream architecture}
    \label{fig:block_diagram}
\end{figure}

Our architecture (Fig. \ref{fig:block_diagram}) is based on a hypothesis of two stream human perception \cite{goodale1992separate}. According to this hypothesis, human vision system consists of two ‘broad’ streams of projection: a 'ventral stream' of projections which is responsible for object recognition, and a 'dorsal stream' that processes the action information. This approach demonstrated good results in computer vision tasks, related to action recognition \cite{simonyan2014two}.

In our approach, we naturally divide the video into two separate components. One component is spatial, and it is represented as a set of independent sequential frames. Another element is temporal, and it relates to changing information on sequential frames. The method was constructed accordingly and has two corresponding signal paths or streams.
 
Using the static information in the form of separate frames is an obvious approach, and it has already been previously used for similar purpose \cite{mascaro2001finger}. It was shown that a spatial stream on its own can demonstrate good results for this task. But, as it will be revealed later in Sect. \ref{sec:results}, the temporal stream may perform better than spatial and dramatically increase the accuracy of estimation.

The overall method is divided into three steps: general image preprocessing, stream processing and stream fusion. Each stream is implemented as a process which consists of two components: preprocessing function and machine learning model. Each stream has a different preprocessing part, but similar machine learning phase. Kalman filter was used to combine information from each stream to get the final estimation of force applied.
 
We rationally assumed that absolute value of a current amount of force can be estimated using spatial stream, and dynamic information of force (the derivative of force applied) can be estimated using temporal stream. It is also assumed, that merging two streams will make the model more accurate and robust compared with each independent stream.

\subsection{General image preprocessing}
\label{subsec:im_preproc}
\begin{figure}[h]
    \centering
    \begin{subfigure}[t]{0.28\linewidth}
        \includegraphics[width=\textwidth]{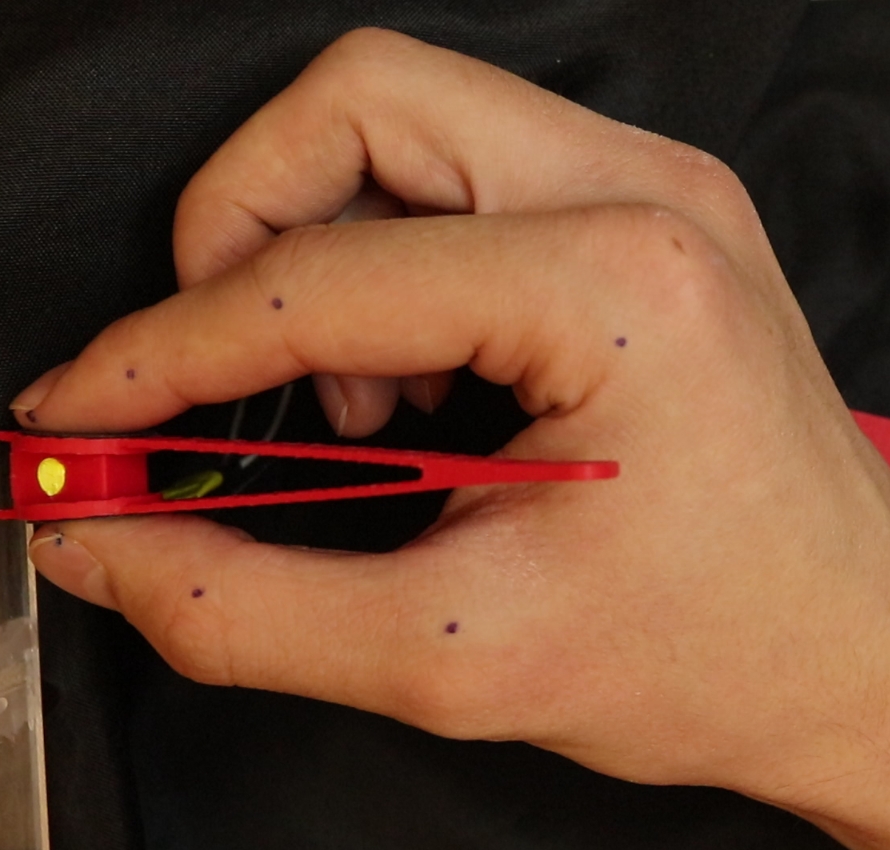}
        \caption{Initial frame of a video}
        \label{fig:initial_frame}
    \end{subfigure}
    \quad
    \begin{subfigure}[t]{0.28\linewidth}
        \includegraphics[width=\textwidth]{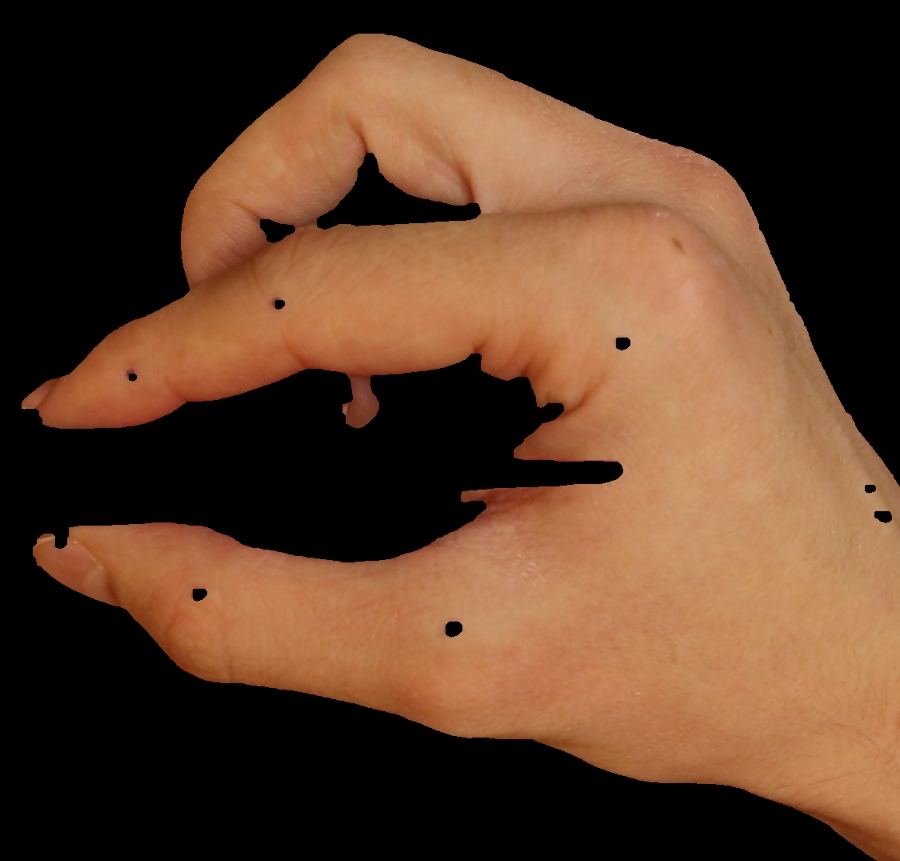}
        \caption{The frame after skin segmentation. Only skin is shown on this frame}
        \label{fig:skin_frame}
    \end{subfigure}
    \quad
    \begin{subfigure}[t]{0.28\linewidth}
        \includegraphics[width=\textwidth]{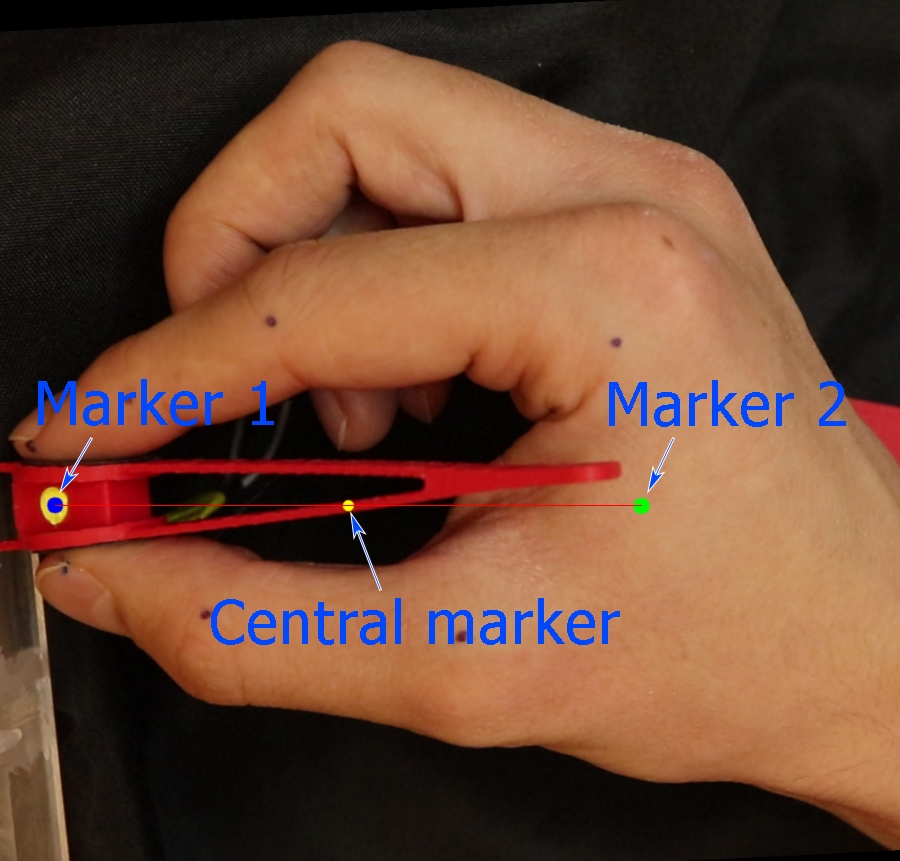}
        \caption{The frame with canceled rotational and translational movements
        }
        \label{fig:markers_final}
    \end{subfigure}
    \caption{Markers parsing and motion canceling}\label{fig:markers}
\end{figure}

General image preprocessing was done for every frame of a video (Fig. \ref{fig:initial_frame}) before feeding it to spatial and temporal streams. The preprocessing phase aimed to mask all the irrelevant parts of a frame and cancel big movements of the camera and a hand. Rotation, translation and scale components of movements were calculated and canceled. This actions allowed to uniformly represent frames independently of small hand position and orientation changes.

During the preprocessing phase the following actions were performed:
\begin{enumerate}
\item Segmentation of skin
\item Marker recognition
\item Scale, translation and rotation calculation, and compensation
\end{enumerate}
 
The segmentation of skin (Fig. \ref{fig:skin_frame}) was done using naive computer vision algorithm. Threshold filter was applied to roughly separate skin from other objects and background. Several iterations of successive morphological operations were applied to the thresholded image to filter out the noise and refine the edge. 

Two markers were detected on each frame to ease image processing. The first marker was applied to a tool itself and had a distinct yellow color. This marker was located between two Force Sensitive Resistors. This marker was segmented using threshold and morphology filters (Fig. \ref{fig:markers_final}). The second marker was calculated as a center of mass of skin region (Fig. \ref{fig:skin_frame}), and for the used grip pattern it was approximately located at the point between Metacarpophalangeal joints of an index finger and a thumb. The central marker was found in the middle of the line, which connected first and second markers (Fig. \ref{fig:markers_final}). Those markers were calculated for rough and fast estimation of frame scale and rotation. 

Using position of marker one and marker two, scale, transition and rotation operations were applied to achieve following conditions:
\begin{itemize}
\item The distance between marker one and the central marker has unit length,
\item Axis X starts at marker one and points toward marker two
\item Marker one has \((0, 0)\) coordinate.
\end{itemize}

\subsection{Spatial stream implementation}
\label{subsec:spatial}

The spatial stream is used to estimate the absolute amount of gripping force applied to an object. The spatial stream may be considered as a robust standalone method. However, as it will be shown later, it does not demonstrate sufficient accuracy on its own. Nevertheless, it is an essential part of the overall method. It allows eliminating accumulated error which may be introduced by integration of spatial stream output.
 
\begin{figure}[h]
    \centering
    \includegraphics[width=0.9\linewidth]{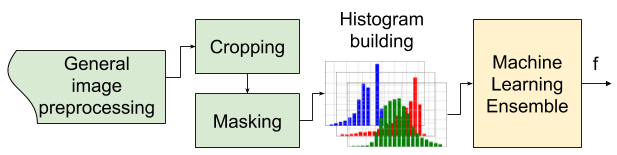}
    \caption{Spatial stream architecture}
    \label{fig:spatial_diagram}
\end{figure}

The overall scheme of the spatial stream is illustrated in Fig. \ref{fig:spatial_diagram}.
The basic idea behind the proposed method is similar to \cite{mascaro2001finger,chen2014estimating}: we exploit the fact that force applied to a fingertip causes skin color change. However, the previously proposed method is based on nail color change. That requires an additional device to be installed on a nail or a nail to be clearly visible by the camera during operation. We mostly concentrate on skin color change instead, as the skin is more likely to be visible.
 
\begin{figure}[h]
    \centering
    \begin{subfigure}[t]{0.3\linewidth}
        \includegraphics[width=\textwidth]{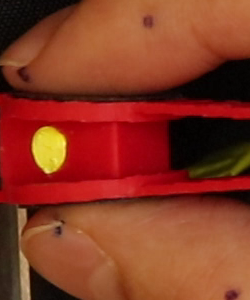}
        \caption{Region of the preprocessed frame with fingertips cropped}
        \label{fig:init_tips}
    \end{subfigure}
    \quad
    \begin{subfigure}[t]{0.3\linewidth}
        \includegraphics[width=\textwidth]{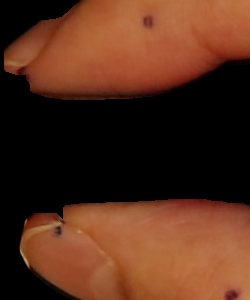}
        \caption{Cropped image with skin mask applied}
        \label{fig:masked_tips}
    \end{subfigure}
    \caption{Fingertips region of a frame}\label{fig:tips}
\end{figure}

Spatial stream preprocessing function uses an output of general image preprocessing algorithm. 
Region with fingertips is cropped (Fig. \ref{fig:init_tips}) from obtained image and skin mask is applied (Fig. \ref{fig:masked_tips}). 
The region of interest is determined as a rectangular region around marker one with constant side lengths width \(l1=0.68\) and height \(l2=0.82\). This length was chosen as it includes only the distal phalanges of an index finger and a thumb. This phalanges primarily stayed immobile during the operation. Moreover, the main skin color change is constrained in the chosen region.

\begin{figure}[h]
    \centering
    \begin{subfigure}[t]{0.7\linewidth}
        \includegraphics[width=\textwidth]{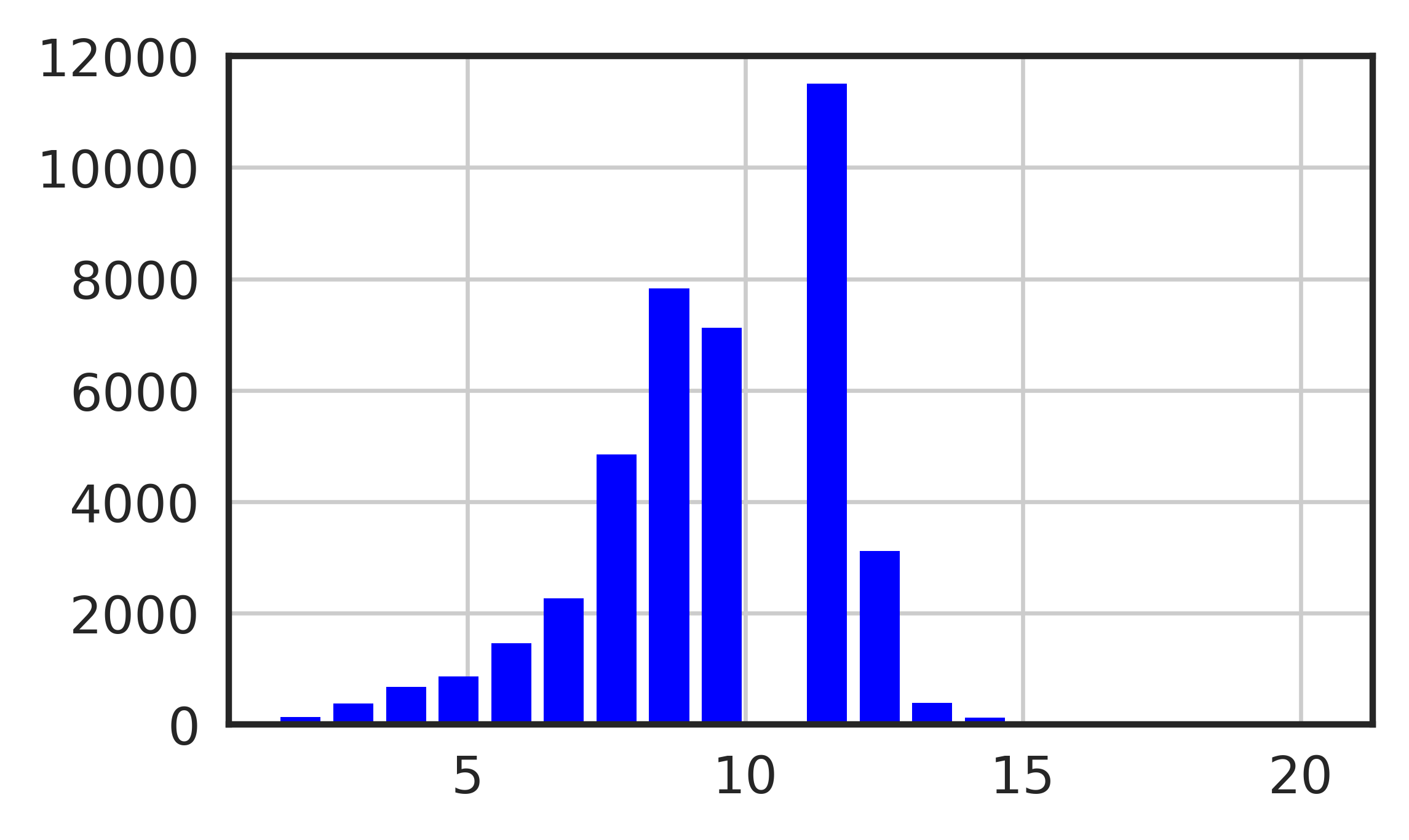}
        \caption{Hue histogram}
        \label{fig:hist_hue}
    \end{subfigure}
    \quad
    \begin{subfigure}[t]{0.7\linewidth}
        \includegraphics[width=\textwidth]{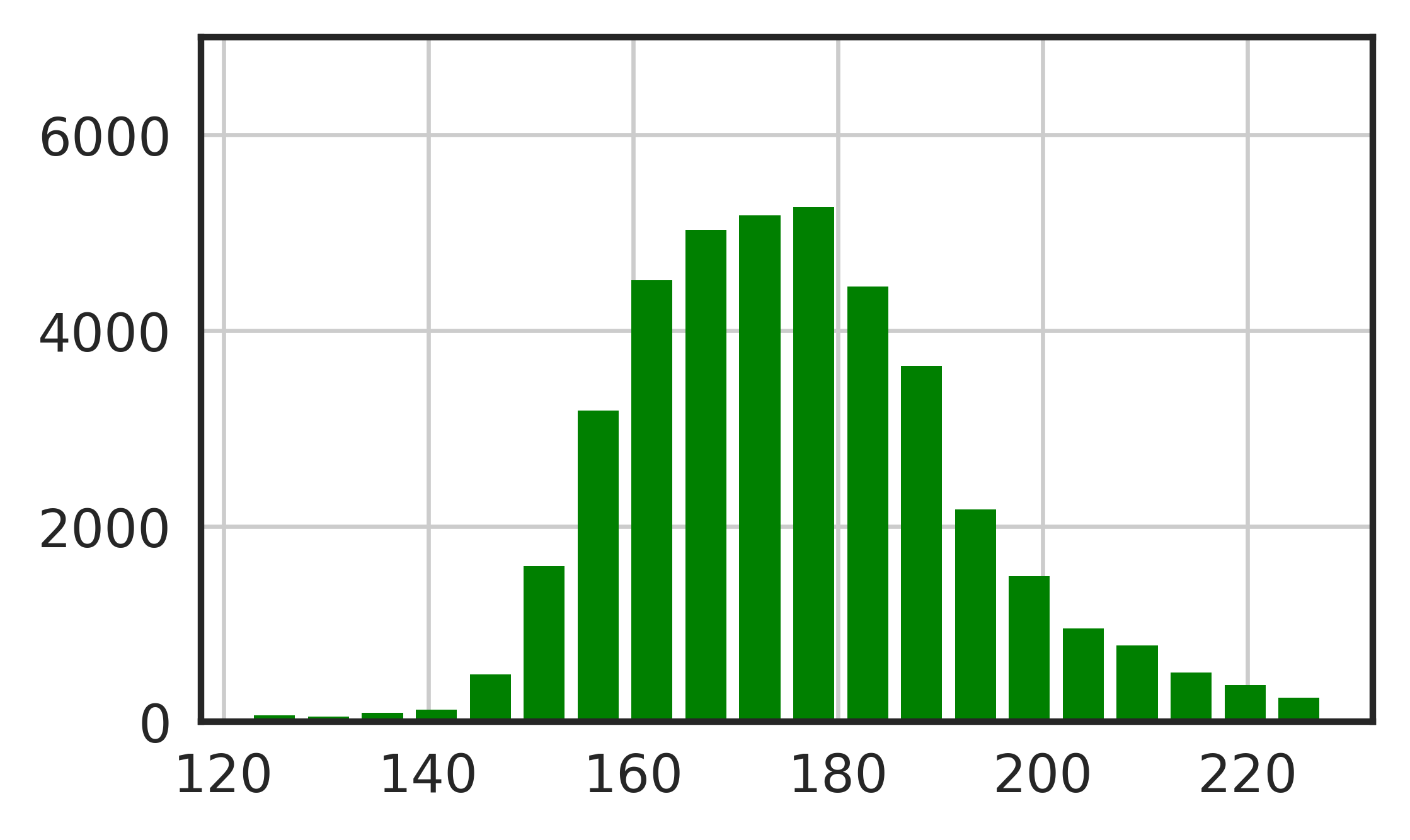}
        \caption{Saturation histogram}
        \label{fig:hist_sat}
    \end{subfigure}
    \quad
    \begin{subfigure}[t]{0.7\linewidth}
        \includegraphics[width=\textwidth]{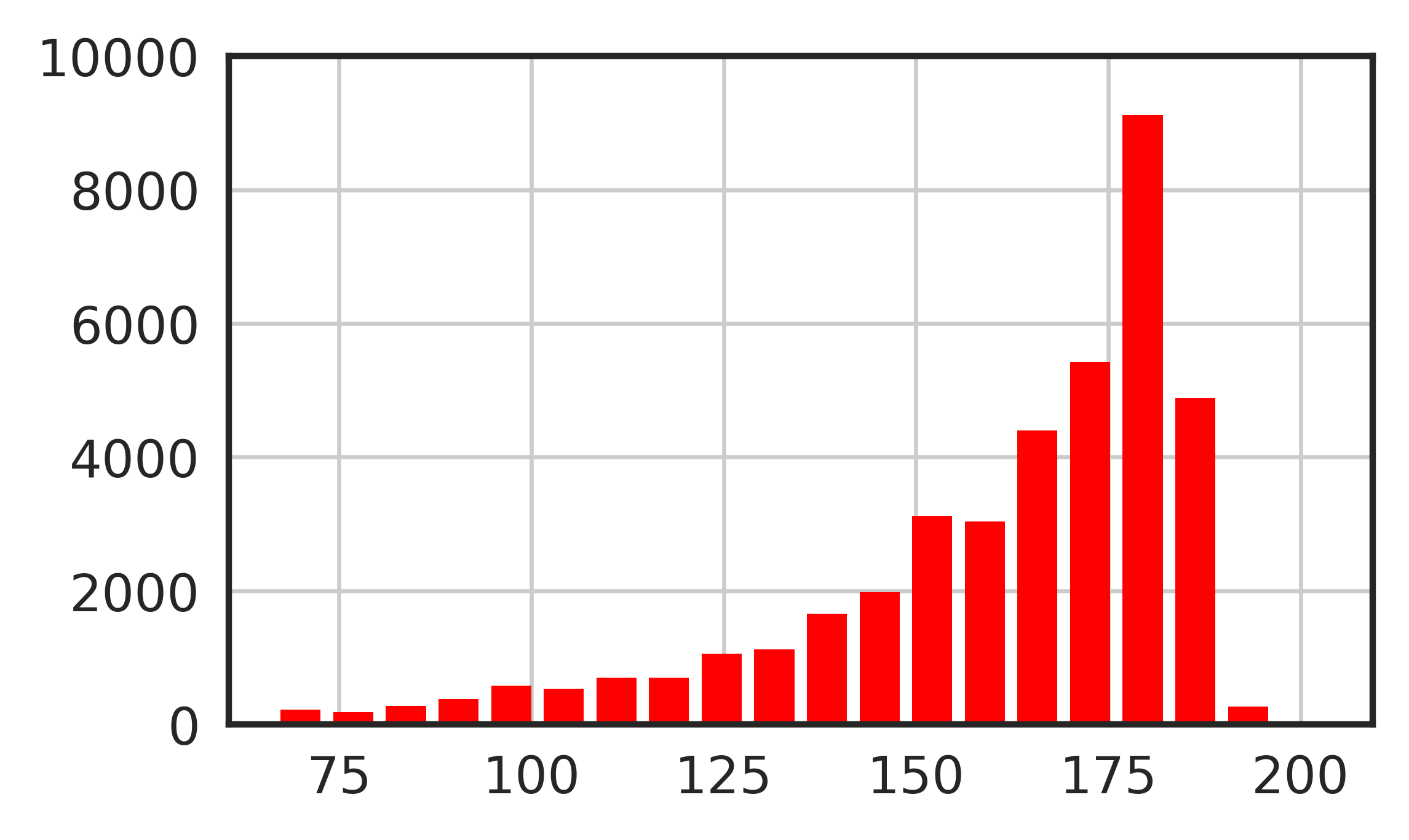}
        \caption{Value histogram}
        \label{fig:hist_val}
    \end{subfigure}
    \caption{Histogram for masked fingertips region}\label{fig:hist_tips}
    \label{fig:tips_hist}
\end{figure}

The relative histogram of this image in HSV space is computed based on cropped and masked image (Fig. \ref{fig:masked_tips}). The histogram is shown in Fig. \ref{fig:tips_hist}. For each separate channel, a histogram was computed. Each histogram has 20 bins. Limits for histograms were chosen in a way to utilize the range of the significant part of histogram more. Combination of three histograms forms 60 features for further processing.

\subsection{Temporal stream implementation}
\label{temporal}

\begin{figure}[h]
    \centering
    \includegraphics[width=0.9\linewidth]{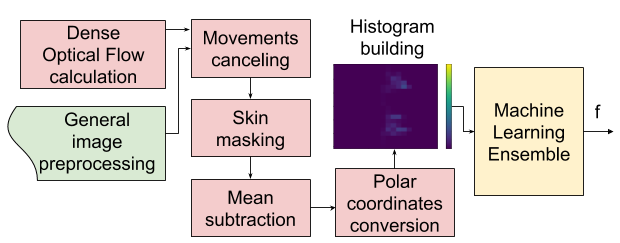}
    \caption{Temporal stream architecture}
    \label{fig:temporal_diagram}
\end{figure}

The temporal stream is used to estimate the first derivative of grip force. This method, as it will be shown later, demonstrates sufficient accuracy, but in some circumstances, it may have an accumulated error due to integration during fusion. The scheme of temporal stream implementation is shown in Fig. \ref{fig:temporal_diagram}. The method is based on dense optical flow \cite{2016_Adarve_RAL} image processing. It utilizes an assumption that optical flow distribution pattern of clenching hand correlates with the gripping force difference.

During the preprocessing phase, dense optical flow of the initial video is calculated using implementation from \cite{pathak2017learning} with the following parameters: number of pyramids \(\alpha = 0.012\), \(ratio = 0.75\), \(minWidth = 20\), \(nOuterFPIterations = 7\), \(nInnerFPIterations = 1\), \(nSORIterations = 30\). The optical flow is further processed. Rotation, translation and scale operations, obtained during the general preprocessing step, are applied to optical flow to cancel big movements and align optical flow respectively. Skin mask from general preprocessing step is used to zero out non-hand movements, such as movements of background objects, movements of the camera. For the masked optical flow matrix mean along all the image for each component is computed. A two-component vector is obtained and subtracted from each non-masked component of optical flow. This step allows minimizing solidary movement of a hand further. The output of this step is referred to as rectified optical flow.

Rectified optical flow vectors are converted to polar coordinate space. Based on vector direction, vector magnitude sign is determined: if vector points towards the middle line, it gets positive sign, otherwise negative. More formally it can be written as (\ref{eq:OF_magn_test}).
\begin{equation}
\begin{cases}
of_m^{new} = -of_m^{old} & \text{if }(of_y-c_y)*of_{\alpha}>0 \\
of_m^{new} = of_m^{old} & \text{if }(of_y-c_y)*of_{\alpha}\leq 0
\end{cases}
\label{eq:OF_magn_test}
\end{equation}
where:
\begin{description}
\item[\(of_m^{new}\)] is the new magnitude of the vector
\item[\(of_m^{old}\)] is the old magnitude of the vector
\item[\(of_y\)] is the y position of the vector
\item[\(of_{\alpha}\)] is the angle of the vector orientation
\item[\(c_y\)] is the y position of the central marker
\end{description}

The motivation is derived from an assumption, that if optical flow vector is pointed toward the middle line, hand clenches and this vector adds up to overall force change. Otherwise, the hand unclenches and the total force change decreases.

\begin{figure}[h]
    \centering
    \begin{subfigure}[t]{0.45\linewidth}
        \includegraphics[width=\textwidth]{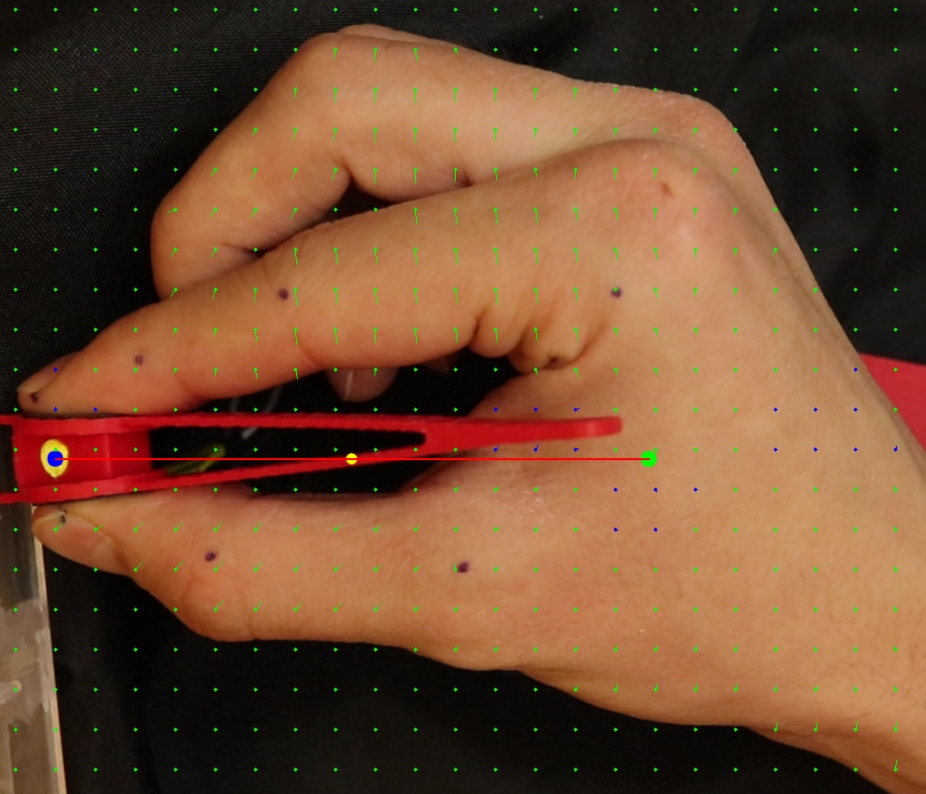}
        \caption{Optical flow of clenching frame}
        \label{fig:OF_clench}
    \end{subfigure}
    \quad
    \begin{subfigure}[t]{0.45\linewidth}
        \includegraphics[width=\textwidth]{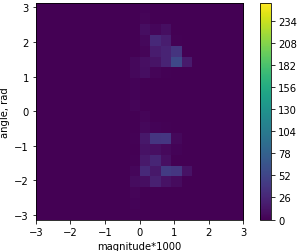}
        \caption{2D histogram of clenching frame}
        \label{fig:hist_clench}
    \end{subfigure}
    \begin{subfigure}[t]{0.45\linewidth}
        \includegraphics[width=\textwidth]{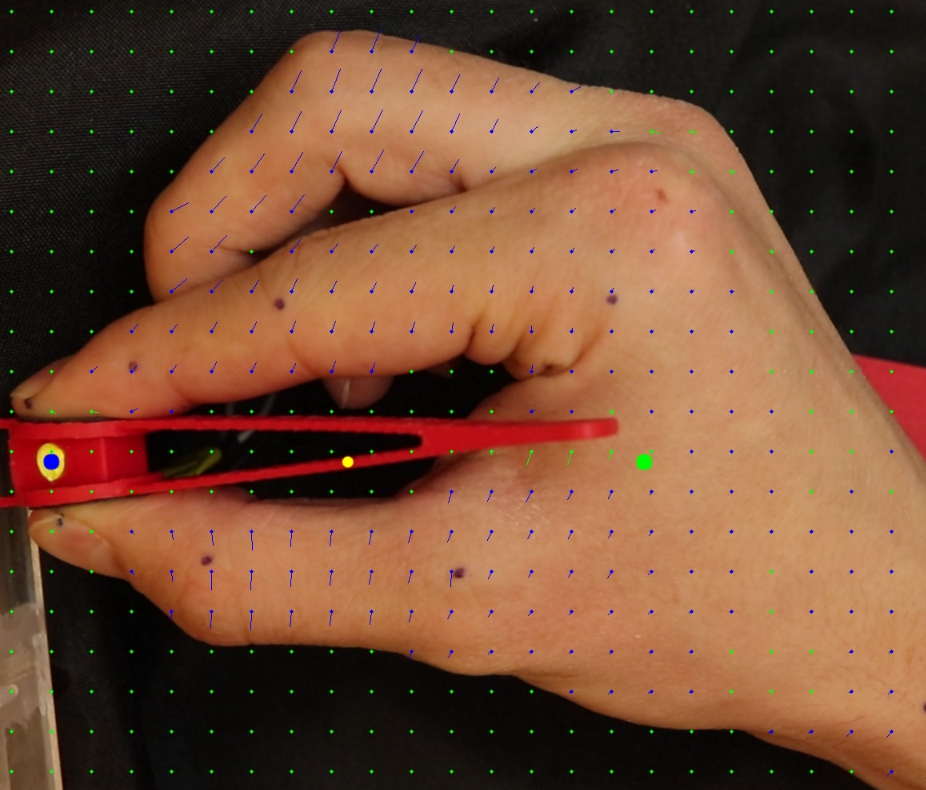}
        \caption{Optical flow of unclenching frame}
        \label{fig:OF_unclench}
    \end{subfigure}
    \quad
    \begin{subfigure}[t]{0.45\linewidth}
        \includegraphics[width=\textwidth]{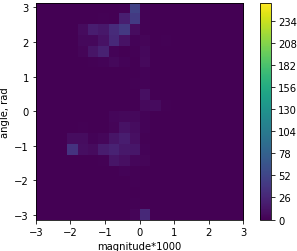}
        \caption{2D histogram of unclenching frame}
        \label{fig:hist_unclench}
    \end{subfigure}
    \caption{Optical flow and histograms for clenching and unclenching hands. Green lines represent optical flow vectors with positive magnitude, blue lines - with negative}\label{fig:OF_hist}
\end{figure}

Optical flow vectors are combined into a two dimensional histogram (Fig. \ref{fig:hist_clench}, \ref{fig:hist_unclench}), with one axis being directions of vectors and the other axis being magnitudes of vector. Histogram limits are chosen in such a way, that \(95\%\) of all the optical flow vectors found in our measurements fall within the limits.

\subsection{Filtering}
\label{subsec:filtering}

Both histograms and discrete difference of force measurements \(\Delta F\) were filtered. For filtering Butterworth low-pass filter was used. Its characteristics are shown in Table \ref{table:filter_params}. The cutoff frequency was chosen to be \(3 Hz\) as in our experiments it was the maximum main frequency with which human could fully controllably clench and unclench test object.

\begin{table}[h]
\caption{Butterworth filter parameters}
\begin{tabular}{l c c c}
Parameter    & Cut-off frequency & Order & Sample rate \\
\hline
Value         & 3Hz                 & 1     & 59.95   \\
\end{tabular}
\label{table:filter_params}
\end{table}

\subsection{Machine learning ensemble train and prediction}
\label{subsec:ml}

Derived histograms were used to train a machine learning ensemble. This step is the same for spatial and temporal streams, but each stream has a separate model. We used auto-sklearn \cite{NIPS2015_5872} to find and train ensembles. For the spatial stream, we used 60 features data points and for temporal 100 features data points.
 
In total the dataset consisted of two 100 sec videos with 59.95 fps frame rate. That gave us 11990 frames in total. From that dataset 8990 points were used for training and 3000 for testing. Sequential sets of frames were selected for testing. Each set consisted of 1500 images or 25 seconds of video. This data has never been seen by model during the training phase. Successive sequences instead of random frames were chosen for two reasons:
Successive frames may be very similar. Thus if train and test sets are chosen randomly, the model can merely remember one frame, and that will allow having a reasonable prediction for a successive frame from the test set.
Having successive sequence ease the visualization of results.
 
As a temporal stream labels average of left and right components of filtered \(\Delta f\) vectors were used. As a spatial stream labels average of left and right components of ground truth force measurements were used.
 

\subsection{Stream fusion}
\label{fusion}

Predictions of the spatial and temporal streams are very noisy. Moreover, the data is redundant: streams predict the variable and the derivative of the variable. Thus to combine results of temporal and spatial streams and filter out noise Kalman filter was used. 

The proposed filter has the magnitude of the force and discrete difference of force as a state vector and as an observations vector (\ref{eq:state}).

\begin{equation}
x_{k}=
\begin{pmatrix}
f_{k} \\
\Delta f_{k}
\end{pmatrix}
\label{eq:state}
\end{equation}

The state transition model addresses the fact, that force magnitude at the next step is the force magnitude at the current step summed up with the discrete difference at the current step. Thus the state-transition matrix can be written as (\ref{eq:transition}):

\begin{equation}
F_{k}=
\begin{pmatrix}
1 & 1 \\
0 & 1
\end{pmatrix}
\label{eq:transition}
\end{equation}

Because the observations vector and the state vector has the same variables, the observation matrix is an identity matrix.

The covariance matrix of the observation noise was composed from ensemble prediction analysis. Thus, since RMSE of spatial stream and RMSE of temporal stream are … and … accordingly, the initial covariance matrix of the observation had the following form (\ref{eq:cov_init}):

\begin{equation}
R_{k}^{init}=
\begin{pmatrix}
50000 & 0 \\
0 & 200
\end{pmatrix}
\label{eq:cov_init}
\end{equation}

The covariance matrix of the process noise was chosen to initially have the form of (\ref{eq:cov_init_proc}). This matrix reflects an assumption that the force on the next step can be calculated with high certainty using the force and discrete difference of force on the previous step. However, the evolution of the derivative of force cannot be predicted.

\begin{equation}
Q_{k}^{init}=
\begin{pmatrix}
1 & 0 \\
0 & 1000
\end{pmatrix}
\label{eq:cov_init_proc}
\end{equation}

Expectation-Maximization algorithm was applied to find better covariance matrix of the observation noise and covariance matrix of the process noise. The final covariance matrix of the observation noise is (\ref{eq:cov_final}):

\begin{equation}
R_{k}^{final}=
\begin{pmatrix}
2523.1 & -15.3 \\
-15.3 & 33.9
\end{pmatrix}
\label{eq:cov_final}
\end{equation}

And the final covariance matrix of the process noise is (\ref{eq:cov_final_proc}):

\begin{equation}
Q_{k}^{final}=
\begin{pmatrix}
1 & -3.5e-3 \\
-3.5e-3 & 22.3
\end{pmatrix}
\label{eq:cov_final_proc}
\end{equation}

\section{Results}
\label{sec:results}

\subsection{Filtering}
\label{subsec:filtering_res}

\begin{figure}[h]
    \centering
    \includegraphics[width=0.9\linewidth]{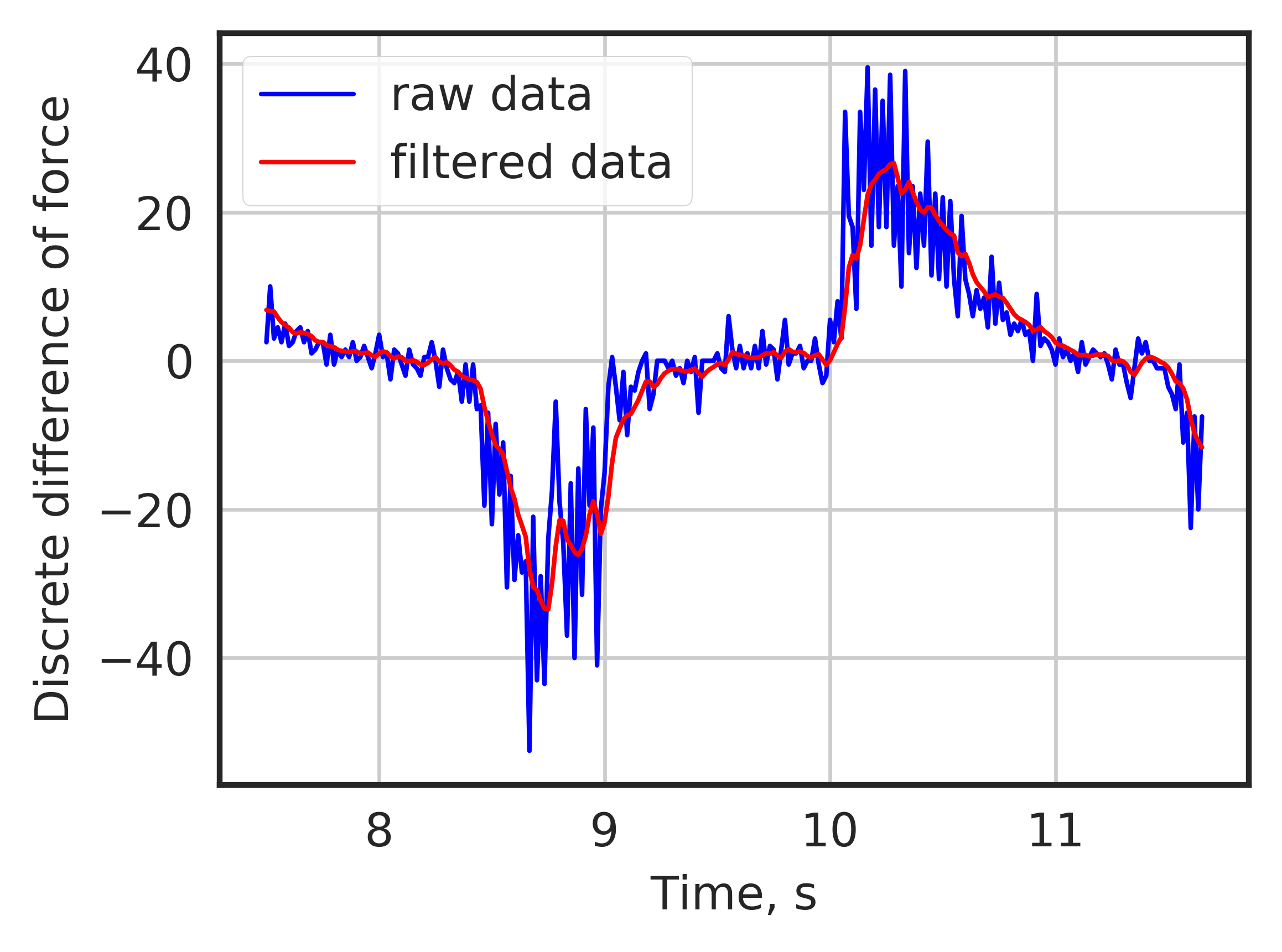}
    \caption{Filtered and not filtered discrete difference of ground truth}
    \label{fig:filter_plot}
\end{figure}

The time plot of filtered and unfiltered discrete difference is shown in Fig. \ref{fig:filter_plot}.

\subsection{Streams and fusion performance}
\label{subsec:rest_res}

\begin{figure}[h]
    \centering
    \begin{subfigure}[t]{0.45\linewidth}
        \includegraphics[width=\textwidth]{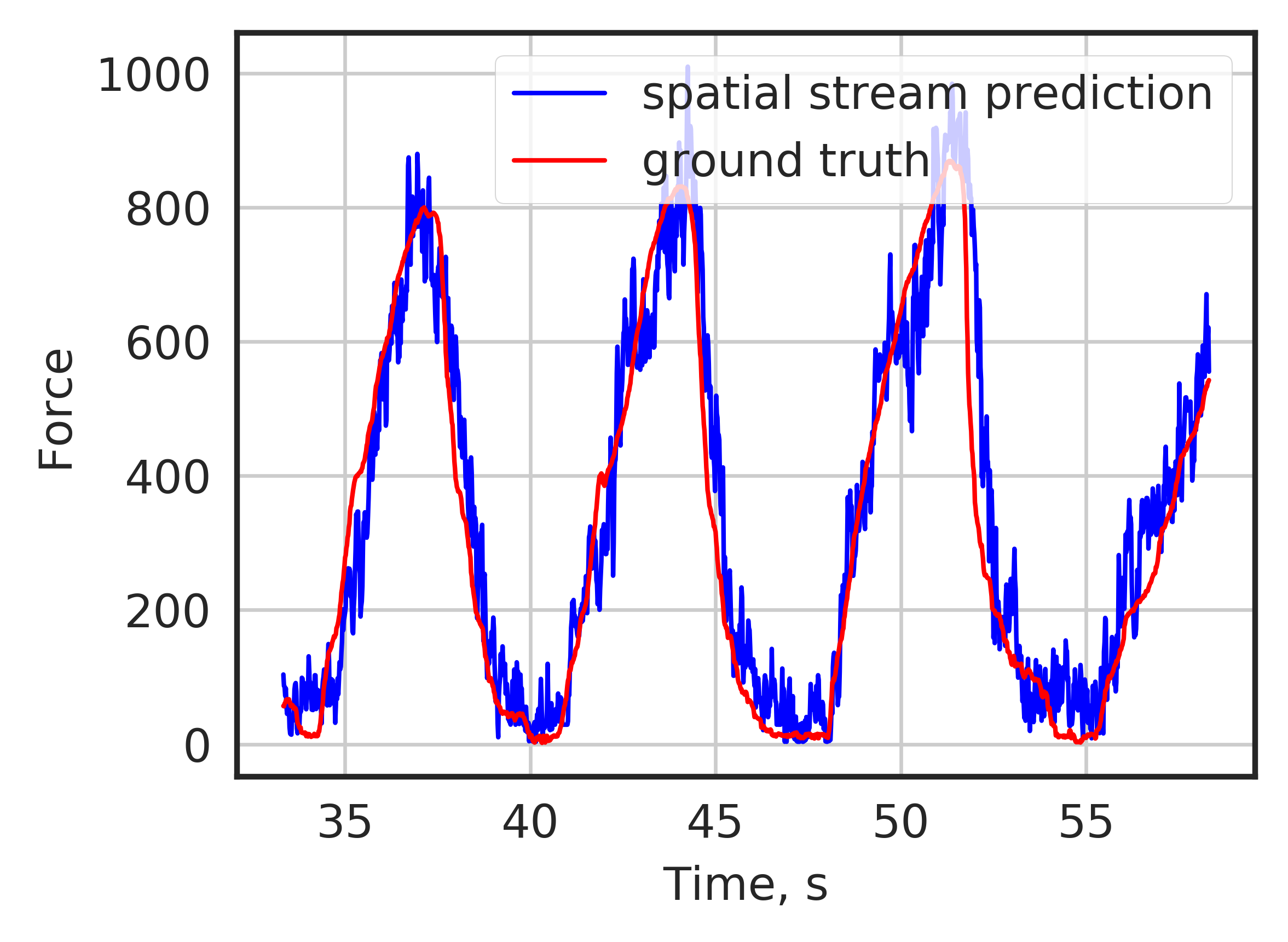}
        \caption{Output of spatial stream and ground truth data for the first video}
        \label{fig:spatial1}
    \end{subfigure}
    \quad
    \begin{subfigure}[t]{0.45\linewidth}
        \includegraphics[width=\textwidth]{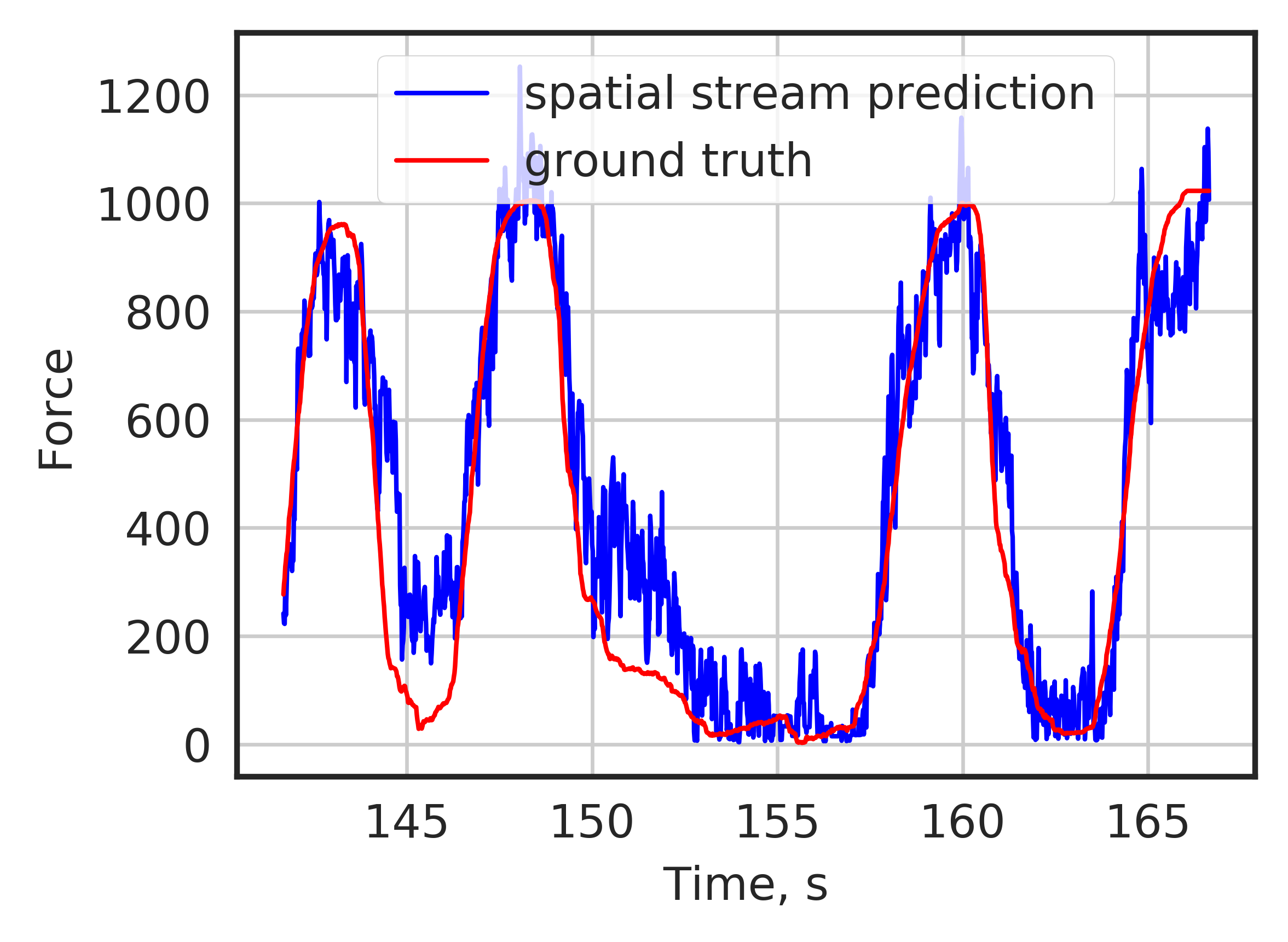}
        \caption{Output of spatial stream and ground truth data for the second video}
        \label{fig:spatial2}
    \end{subfigure}
    \begin{subfigure}[t]{0.45\linewidth}
        \includegraphics[width=\textwidth]{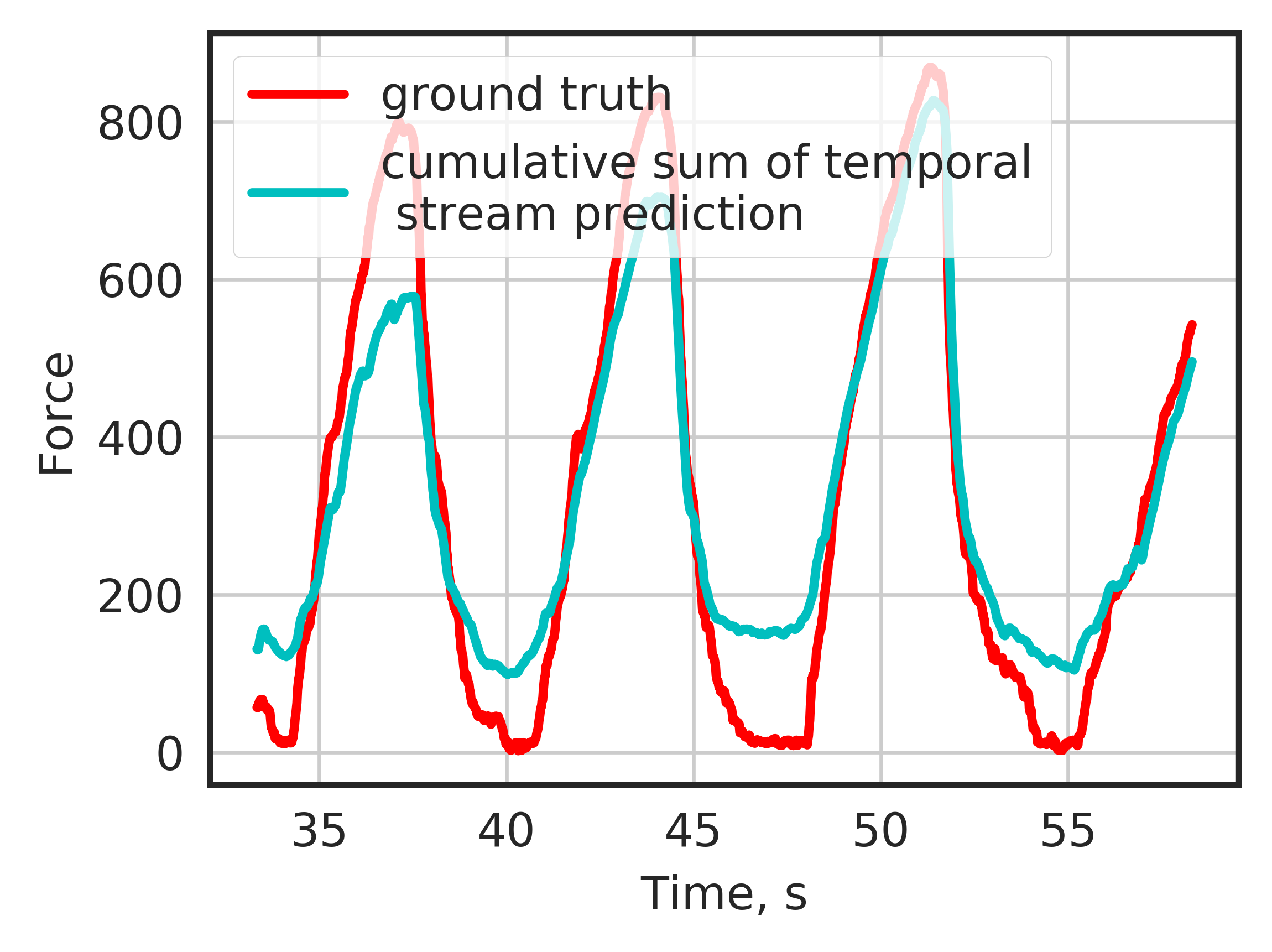}
        \caption{Cumulative sum of the output of temporal stream and ground truth data for the first video}
        \label{fig:temporal1}
    \end{subfigure}
    \quad
    \begin{subfigure}[t]{0.45\linewidth}
        \includegraphics[width=\textwidth]{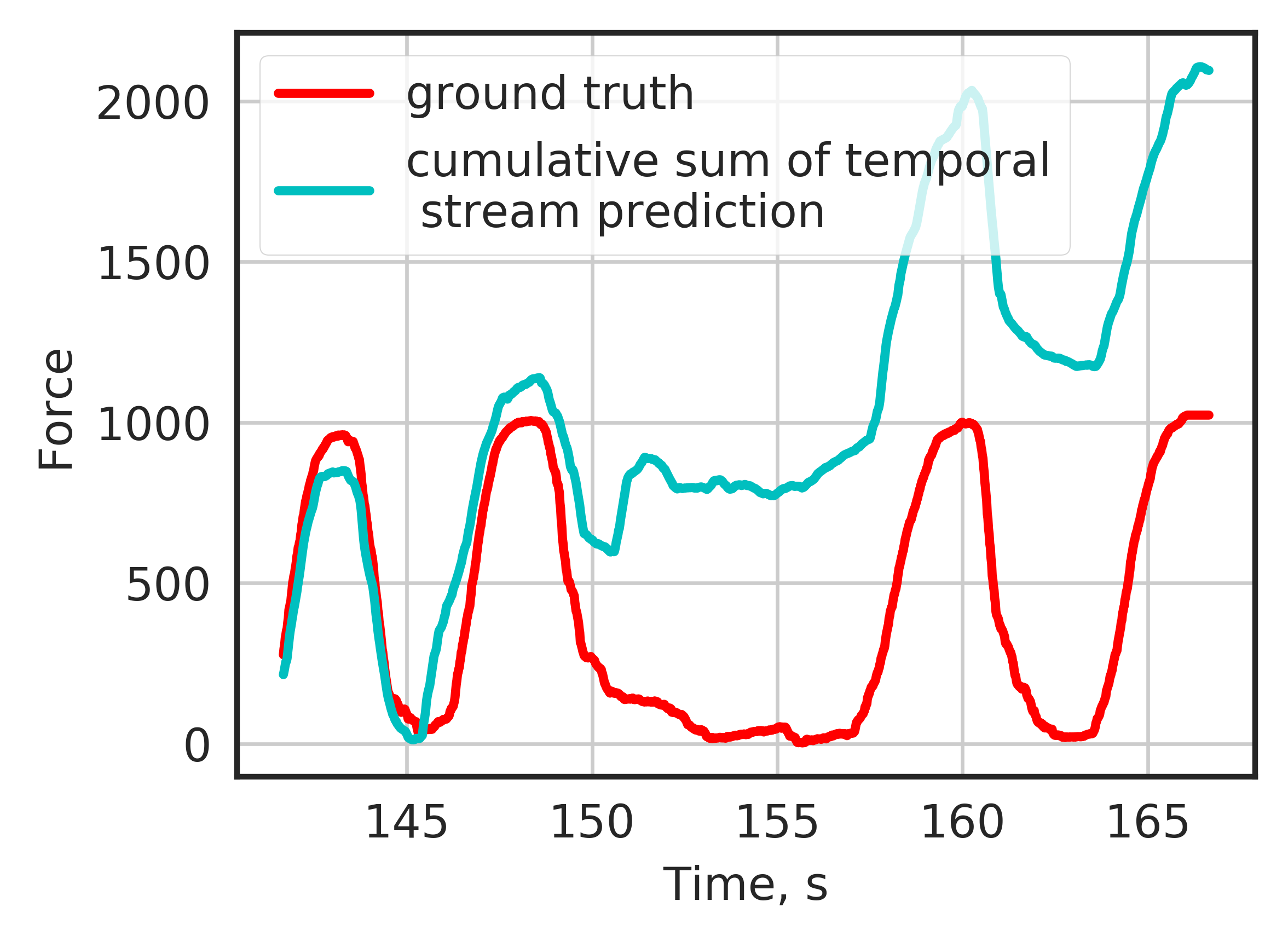}
        \caption{Cumulative sum of the output of temporal stream and ground truth data for the second video}
        \label{fig:temporal2}
    \end{subfigure}
    \begin{subfigure}[t]{0.45\linewidth}
        \includegraphics[width=\textwidth]{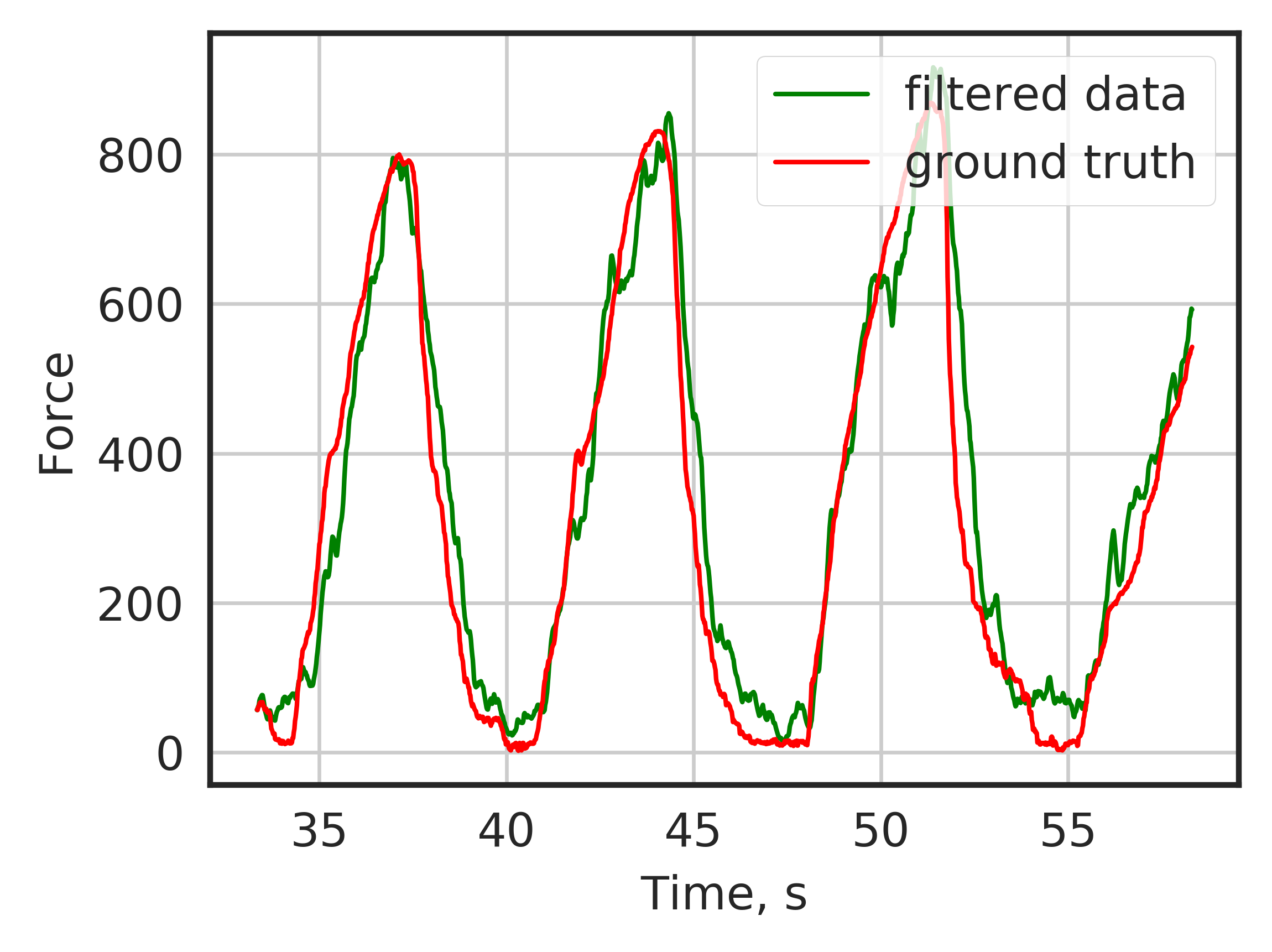}
        \caption{Output of Kalman filter and ground truth data for the first video}
        \label{fig:filtered1}
    \end{subfigure}
    \quad
    \begin{subfigure}[t]{0.45\linewidth}
        \includegraphics[width=\textwidth]{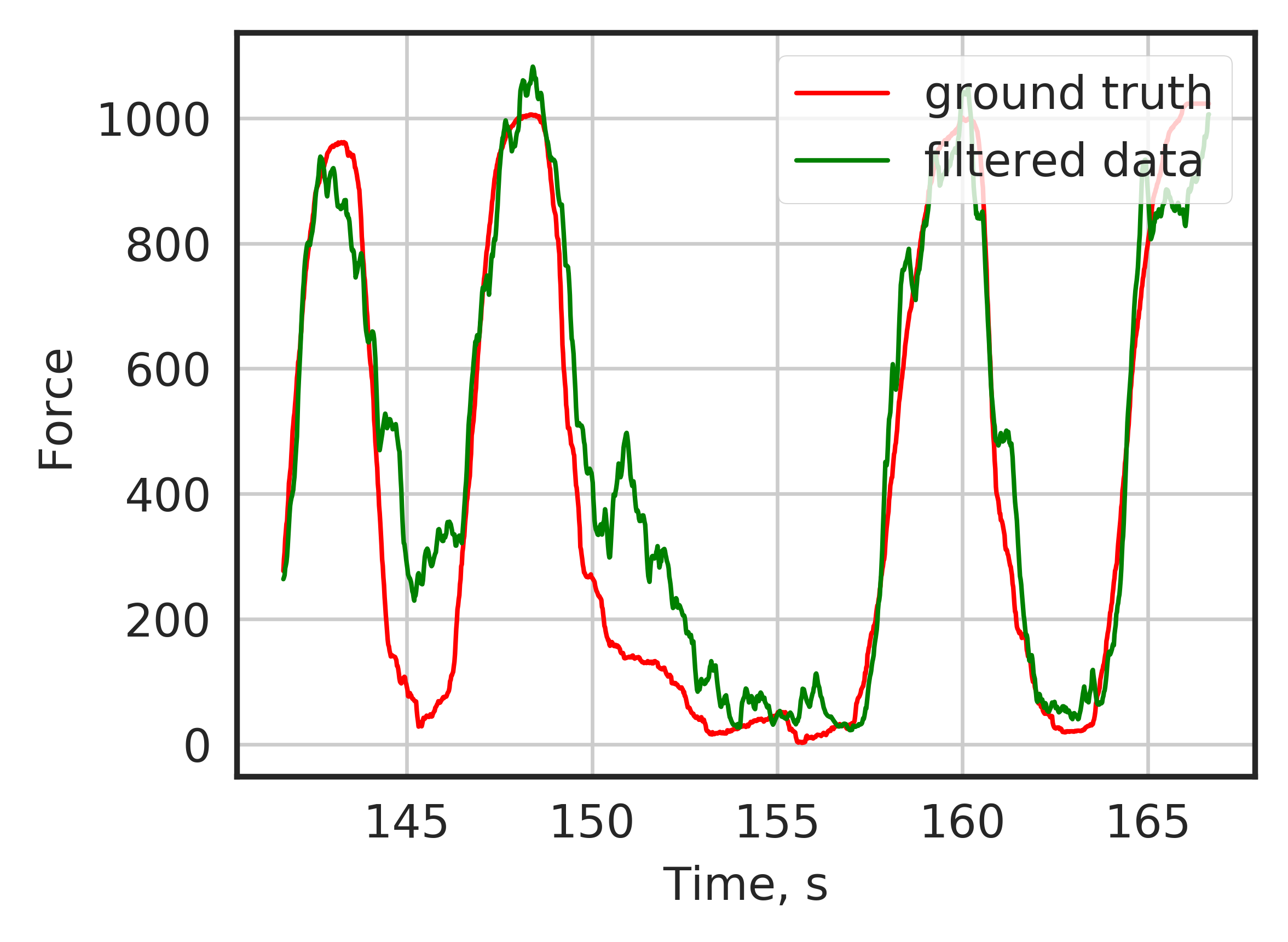}
        \caption{Output of Kalman filter and ground truth data for the second video}
        \label{fig:filtered2}
    \end{subfigure}
    \caption{Single stream and overall algorithm outputs time plots}\label{fig:outputs}
    \label{fig:results}
\end{figure}

The time plot of ground truth values, result of the algorithm and single stream results are shown in Fig. \ref{fig:results}. The results of the temporal stream, spatial stream, and overall result were close for the first video. However temporal stream cumulative sum produced an invalid result in the second case, posing high accumulated error. Comparison of performance of each stream and overall algorithm for both cases are shown in the Table \ref{table:results}. The RMSE score of Kalman filter output was the best compared to the score of every single stream in both cases. The RMSE score of the temporal stream was slightly worse for the first video compared to the overall algorithm and spatial stream. The score of the spatial stream was slightly less than the score of the overall algorithm.

\begin{table}[h]
\caption{Each stream and overall results comparison}
\begin{tabular}{l c c c c}
\hline
{}                & \multicolumn{2}{c}{First video} & \multicolumn{2}{c}{Second video} \\
{}                    & MSE             & r2                & MSE                & r2  \\
\hline
Spatial stream         & 7350.21        & 0.905                & 17597.73            & 0.878   \\
Temporal stream     & 8331.59        & 0.892                & 567813.82            & -2.931 \\
Fused                  & 5795.25        & 0.925                & 15884.96            & 0.890  \\
\hline
\end{tabular}
\label{table:results}
\end{table}

\begin{figure}[h]
    \centering
    \begin{subfigure}[t]{0.4\linewidth}
        \includegraphics[width=\textwidth]{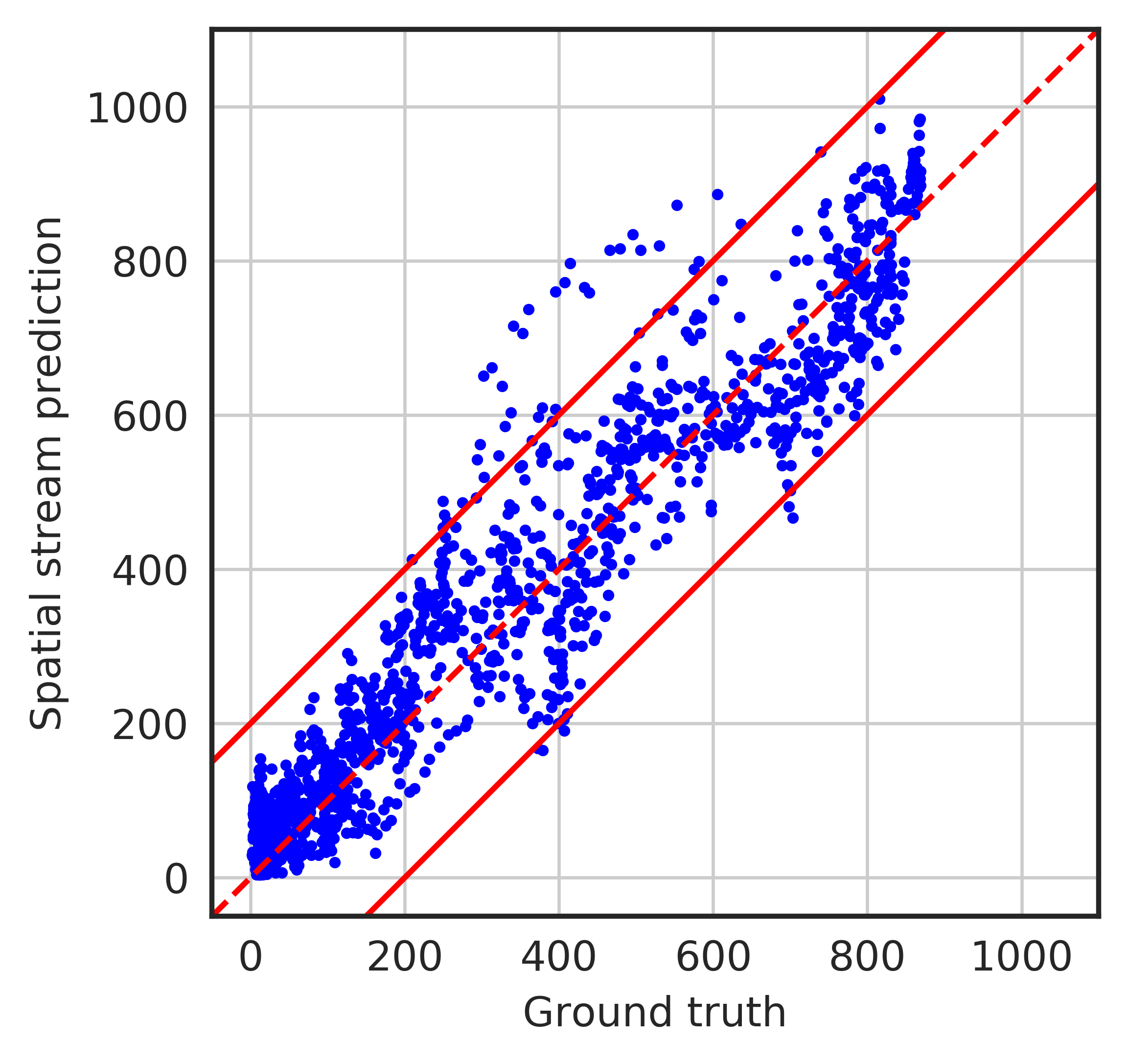}
        \caption{Output of spatial stream and ground truth data for the first video}
        \label{fig:spatial1_cp}
    \end{subfigure}
    \begin{subfigure}[t]{0.4\linewidth}
        \includegraphics[width=\textwidth]{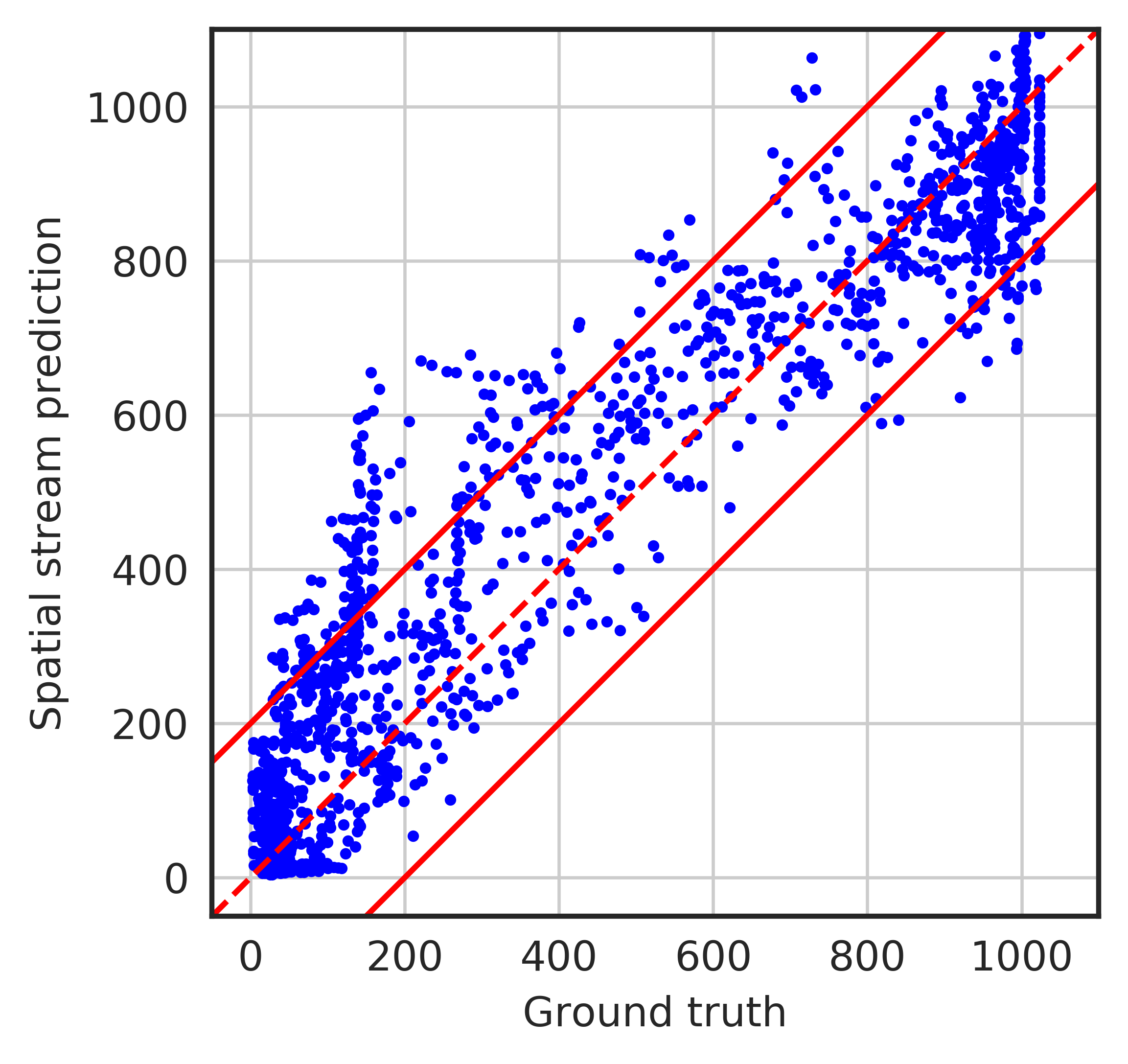}
        \caption{Output of spatial stream and ground truth data for the second video}
        \label{fig:spatial2_cp}
    \end{subfigure}
    \begin{subfigure}[t]{0.4\linewidth}
        \includegraphics[width=\textwidth]{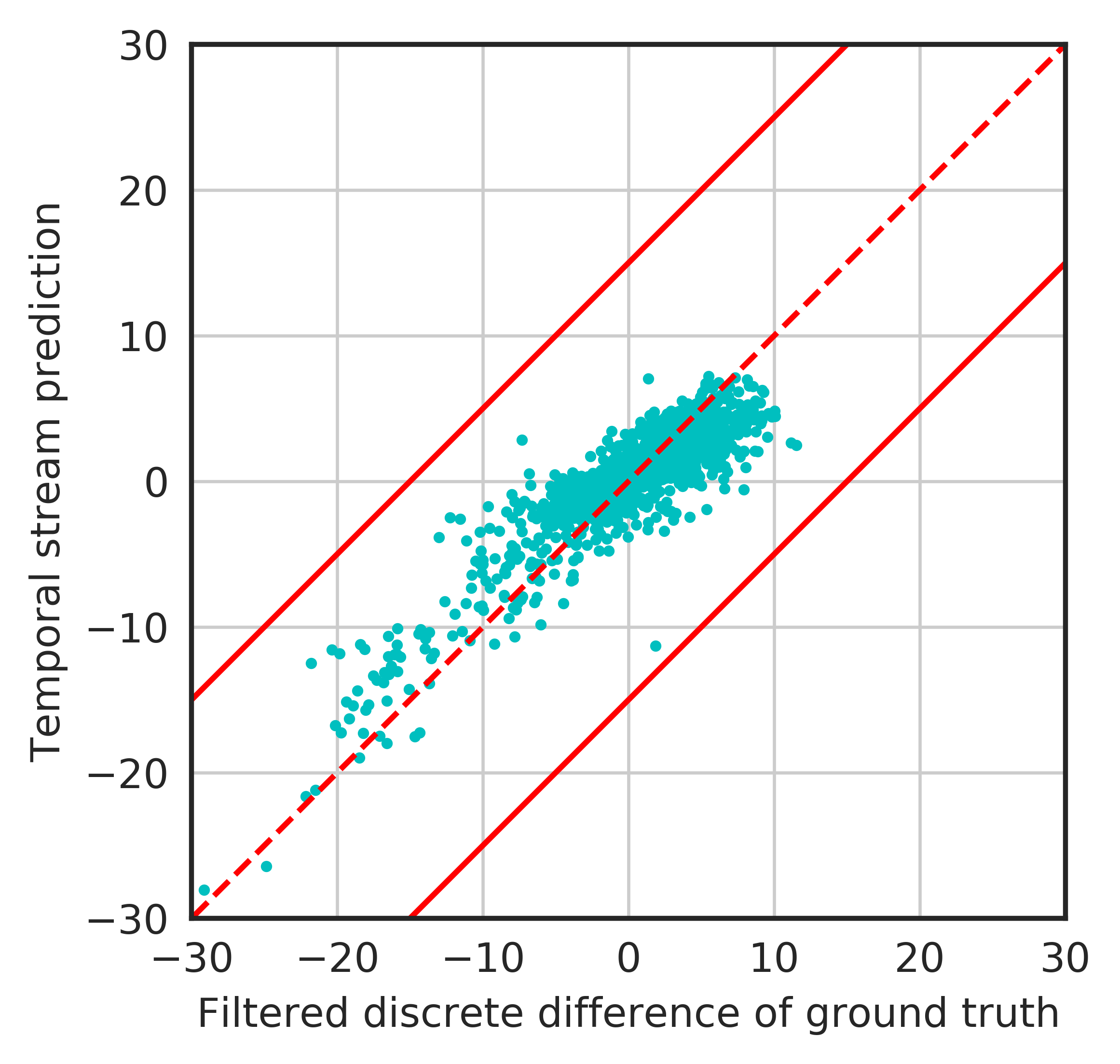}
        \caption{Output of temporal stream and discrete difference of ground truth data for the first video}
        \label{fig:temporal1_cp}
    \end{subfigure}
    \begin{subfigure}[t]{0.4\linewidth}
        \includegraphics[width=\textwidth]{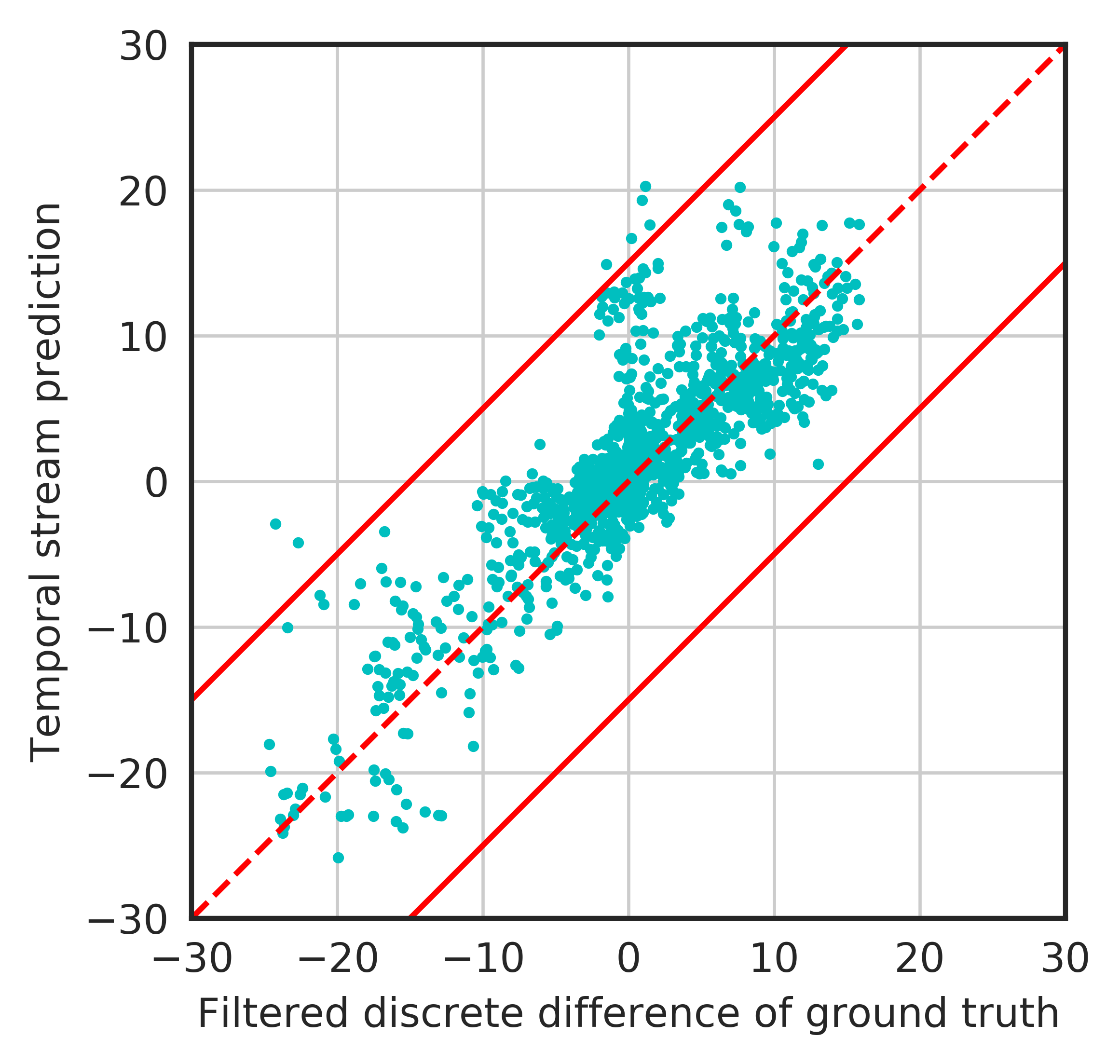}
        \caption{Output of temporal stream and discrete difference of ground truth data for the second video}
        \label{fig:temporal2_cp}
    \end{subfigure}
    \begin{subfigure}[t]{0.4\linewidth}
        \includegraphics[width=\textwidth]{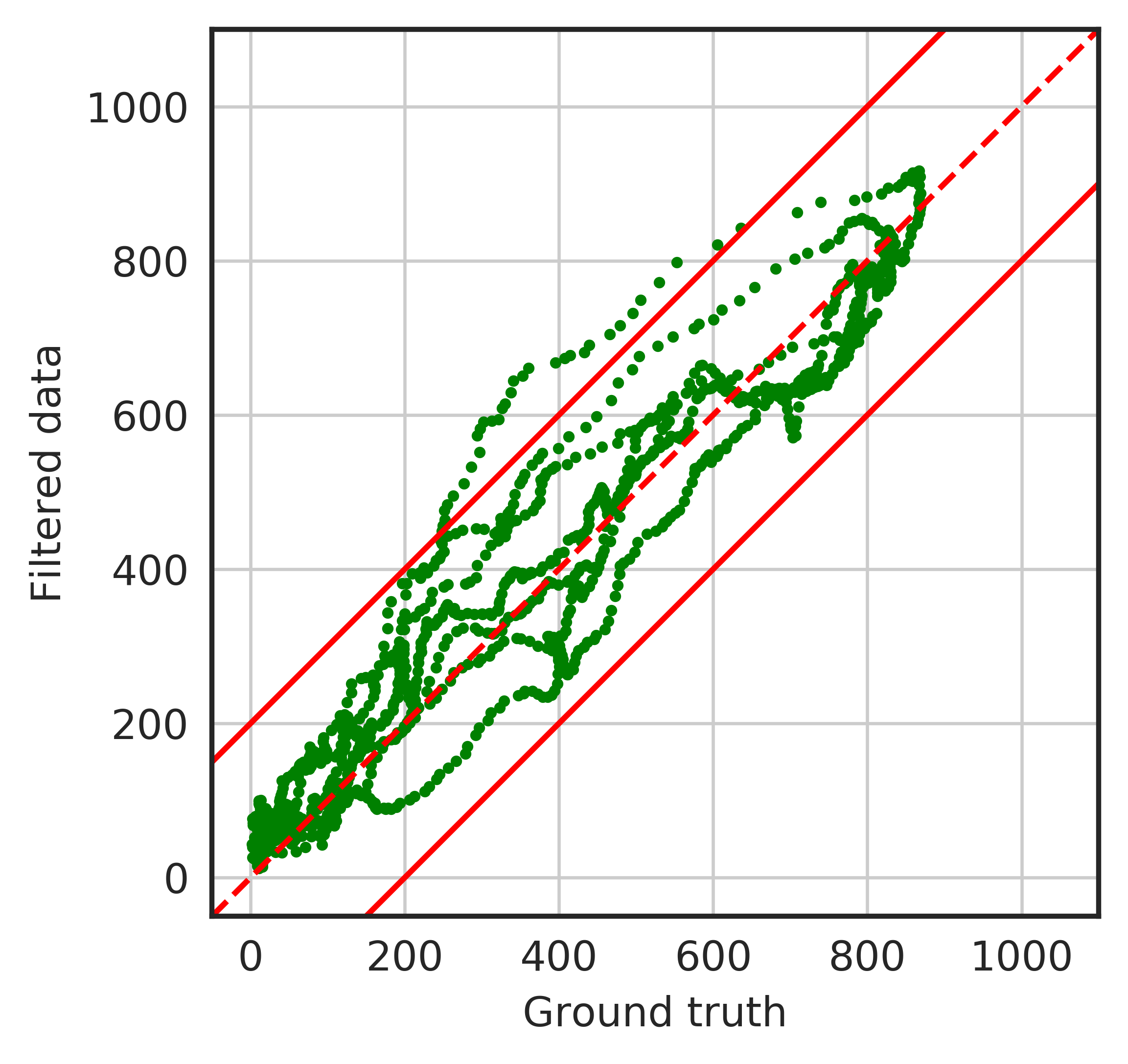}
        \caption{Output of Kalman filter and ground truth data for the first video}
        \label{fig:filtered1_cp}
    \end{subfigure}
    \begin{subfigure}[t]{0.4\linewidth}
        \includegraphics[width=\textwidth]{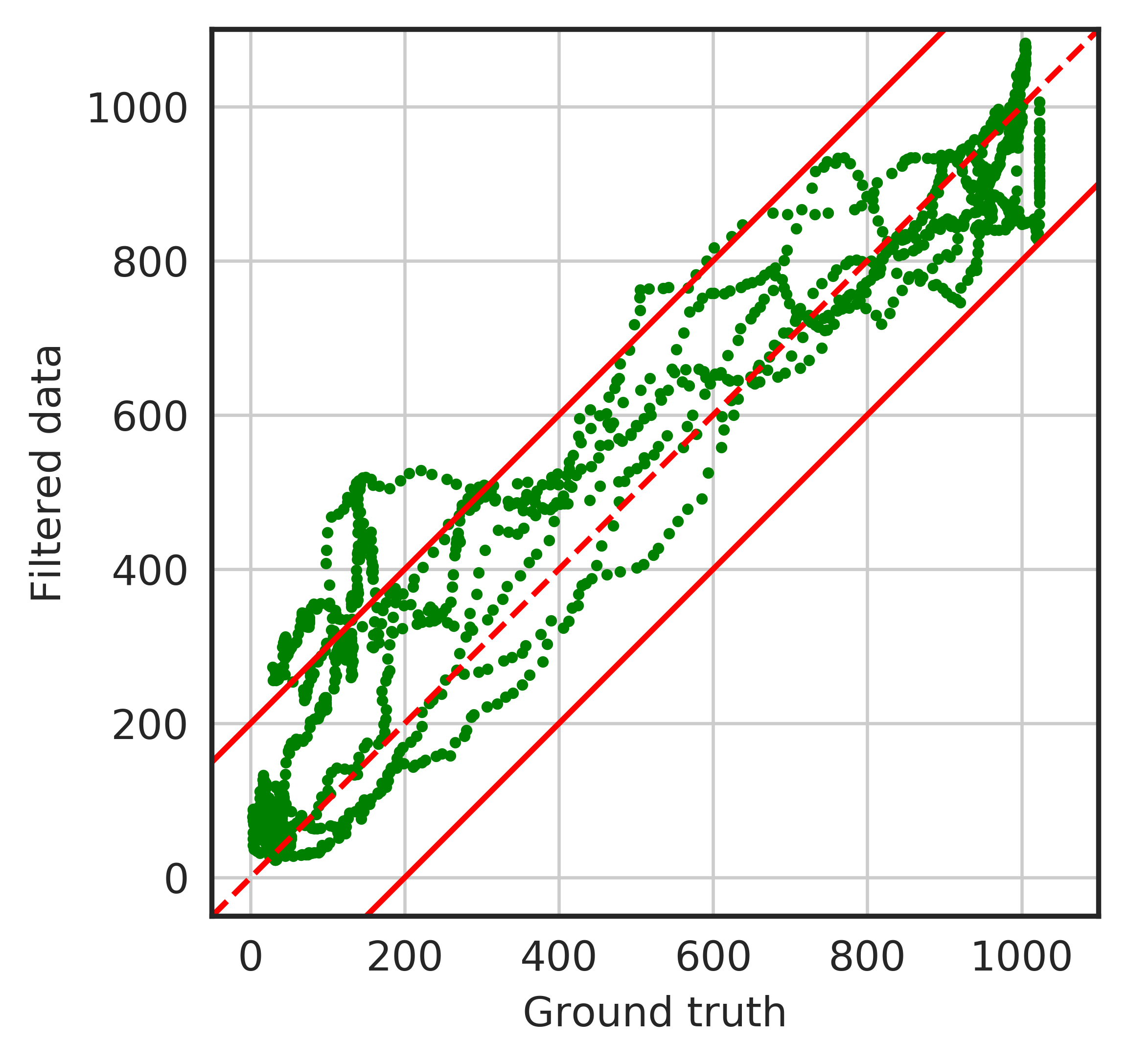}
        \caption{Output of Kalman filter and ground truth data for the second video}
        \label{fig:filtered2_cp}
    \end{subfigure}
    \caption{Single stream and overall algorithm outputs comparison plots}\label{fig:outputs_cp}
\end{figure}

The temporal stream demonstrated reduced performance in the measurements close to zero compared to other measurements in the second video (Fig. \ref{fig:temporal2_cp}). However the spatial stream did not have significant change of performance dependence on the magnitude of the true value (Fig. \ref{fig:spatial1_cp}, \ref{fig:spatial2_cp}). Filtered value did not have notable performance dependence on the magnitude of value either (Fig. \ref{fig:filtered1_cp}, \ref{fig:filtered2_cp})

\section{Discussion and Conclusions}
\label{sec:conclusions}

In this article, we proposed a two-stream algorithm for estimation of gripping force from the video. As was expected, temporal stream based on optical flow showed almost as good results as the overall method, which correlates with the conclusion of \cite{simonyan2014two}. That is a very interesting result, as this method did not directly use color information of the hand, only the information on the movements of the hand. This result may be due to a very low sensitivity of the method itself to the lighting condition. However, the method demonstrated reduced performance when slow changes of the force took place. The possible reason is that in this case, overall noise may be indistinguishable from the actual measurement. Another possible reason is the small amount of training data.

However, the precision of spatial stream itself was less than expected. The cause of that may be the probable high sensitivity of the method to the lighting conditions. The delay between the impact and the actual change of the skin color may also be significant. It is also probable that the method is sensitive to uncompensated angular movements of the hand.

Combination of two streams even with the straightforward Kalman filter improved the result of each separate stream. It also expectedly made the method more robust and less sensitive to the current output value. 

However, the proposed method in its current implementation has several significant limitations. Firstly it was tested on a limited amount of data. Overall 200 seconds of videos were used to train and test the algorithm. The more extensive dataset with the wide variety of humans and test object forms should be used in the future for training and testing the method. Secondly, a relatively weak model was used for predictions. In future works, we plan to strengthen the model against lighting condition changes, hand orientation changes, camera and hand movements. Hand keypoint recognition methods \cite{zb2017hand}, \cite{simon2017hand} for rotation, translation, and scale estimation and canceling will also be considered.

Changing the naive CV method for skin segmentation to CNN based approaches \cite{wei2016convolutional} may also improve the results. Other adaptive \cite{dadgostar2006adaptive}, and more sophisticated \cite{zuo2017combining} methods also may be used for this purpose. Use of these methods may increase the robustness of the system for real-world applications.

Overall, the noninvasive approach to measure the grasp position and force from video is presented. The performance of the algorithm reaches RMSE $\approx 2 N$, which is enough for its application in learning forces of action in automated smartphone disassembly and other robotic applications of similar scale. Application of such algorithm to the available online videos can produce a great dataset of actions, annotated not only with the spatial trajectories but also with required characteristics of the forces applied, which can be naturally executed by modern robots.

\balance
\bibliographystyle{IEEEtran}
\bibliography{IEEEabrv,SkolTech}


\pagebreak
\pagebreak

\end{document}

%% file: ieeepreamble.tex
\usepackage{microtype}
\usepackage{cmap} 

\usepackage{graphics} 
\usepackage{epsfig} 
\usepackage{mathptmx} 
\usepackage{times} 
\usepackage{amsmath} 
\usepackage{amssymb}  

\usepackage{xspace} 

\usepackage{graphicx}  
\graphicspath{{./pics/}} 
\DeclareGraphicsExtensions{%
    .png,.PNG,%
    .pdf,.PDF,%
    .jpg,.mps,.jpeg,.jbig2,.jb2,.JPG,.JPEG,.JBIG2,.JB2}

\usepackage{upgreek} 

\usepackage[nodisplayskipstretch]{setspace}

\usepackage[font=footnotesize]{caption}
\usepackage[font=footnotesize]{subcaption}

\usepackage{bookmark}

\newcommand{\exTODO}[1]{}

\usepackage{listings}
\usepackage{color}